\documentclass{article}

\usepackage[nonatbib,preprint]{neurips_2023}

\usepackage[utf8]{inputenc} 
\usepackage[T1]{fontenc}    
\usepackage{hyperref}       
\usepackage{url}            
\usepackage{booktabs}       
\usepackage{amsfonts}       
\usepackage{nicefrac}       
\usepackage{microtype}      
\usepackage{xcolor}         
\usepackage{tabularx}
\usepackage{wrapfig}

\usepackage{microtype}
\usepackage{graphicx}
\usepackage{caption}
\usepackage{subcaption}

\usepackage{amsmath,amsthm,amssymb}
\usepackage{bm}
\usepackage{amsfonts}

\usepackage{algorithm}
\usepackage{algorithmic}

\title{\textit{AdaSelection}: Accelerating Deep Learning Training through Data Subsampling}

\author{%
	Minghe Zhang\thanks{Work done as an intern at Amazon}\\
	Georgia Institute of Technology\\
	\texttt{minghe\_zhang@gatech.edu} \\
	\And
	Chaosheng Dong\thanks{Corresponding author} \\
	Amazon \\
	\texttt{chaosd@amazon.com}  \\
	\And
	Jinmiao Fu\\
	Amazon \\
	\texttt{jinmiaof@amazon.com}\\
	\And
	Tianchen Zhou\textsuperscript{*}\\
	The Ohio State University\\
	\texttt{zhou.2220@osu.edu}\\
	\And
	Jia Liang\textsuperscript{*}\\
	Stanford University\\
	\texttt{Jialiang@stanford.edu}\\
	\And
	Jia Liu\\
	Amazon \\
	\texttt{hliujia@amazon.com}\\
	\And
	Bo Liu\\
	Amazon \\
	\texttt{boliucs@amazon.com}\\
	\And
	Michinari Momma\\
	Amazon \\
	\texttt{michi@amazon.com}\\
	\And
	Bryan Wang\\
	Amazon \\
	\texttt{brywan@amazon.com}\\
	\And
	Yan Gao\\
	Amazon \\
	\texttt{yanngao@amazon.com}\\
	\And
	Yi Sun\\
	Amazon \\
	\texttt{yisun@amazon.com}\\
}

\begin{document}

	\maketitle

	\begin{abstract}
		In this paper, we introduce \textit{AdaSelection}, an adaptive sub-sampling method to identify the most informative sub-samples within each minibatch to speed up the training of large-scale deep learning models without sacrificing model performance. Our method is able to flexibly combines an arbitrary number of baseline sub-sampling methods incorporating the method-level importance and intra-method sample-level importance at each iteration. The standard practice of ad-hoc sampling often leads to continuous training with vast amounts of data from production environments. To improve the selection of data instances during forward and backward passes, we propose recording a constant amount of information per instance from these passes. We demonstrate the effectiveness of our method by testing it across various types of inputs and tasks, including the classification tasks on both image and language datasets, as well as regression tasks. Compared with industry-standard baselines, \textit{AdaSelection} consistently displays superior performance.
	\end{abstract}

	\section{Introduction}
	
	Large deep learning models (DNNs) trained with data of massive volume continues to push the state-of-the-art across various domains. However, reducing the time and cost incurred remains a challenging problem. It can take days or weeks to train a single image classification model using conventional optimizers such as minibatch Stochastic Gradient Descent due to the high complexity of most model backbones. For example, the ImageNet contests have seen the parameter size of Convolutional Neural Networks (CNN) increase to over $n^{9}$. The gradient computation using back-propagation is the most time-consuming component during the training process. It has been shown that DNNs tend to learn simple patterns first \cite{Arpit2017ACL}, therefore training using these samples at later epochs brings little to no incremental value while being unnecessarily time-consuming. This motivates us to design a framework to identify the most informative sub-samples within each mini-batch during model training.
	
	Existing sub-sampling works mainly formulate the problem with two approaches. One approach is to learn the data importance score per sample and sub-sample the data with probability proportional to the importance weights, such as Big Loss \cite{jiang2019accelerating}, Small Loss \cite{shah2020choosing}, RHO-loss \cite{mindermann2022prioritized}, gradient norm \cite{katharopoulos2018not}, etc. The other approach identifies the best sub-samples in an optimization manner. For example, \cite{dong2021one} casts the sample selection as a Mixed Integer Programming (MIP) problem and solves it with an optimization solver. The optimization problem can also be approximated and solved more efficiently using the Bayesian approach \cite{sener2017active} or Frank-Wolfe approach \cite{campbell2018bayesian}. However, none of these methods provide a clear analysis on the trade-off among training time, sampling rate and accuracy. Additionally, these methods have a fixed policy for each iteration/epoch, which do not adapt to distribution shifts in the training batch. Furthermore, none of the existing methods outperform others consistently across all tasks (classification and regression), therefore it's still unclear to users which method to choose when facing a new task.
	
	To address these drawbacks, we propose \textit{AdaSelection}, an algorithm that adaptively combines different baseline methods to select the top $k\%$ most informative sub-samples at each iteration by leveraging intra-method sample-importance and the inter-method method-importance. Our approach can significantly reduces the training time with only a marginal drop in model performance. We demonstrate our algorithm's SOTA performance across different public datasets with various types of inputs and tasks, including three image classification tasks (Cifar10, Cifar100 and SVHN), two regression tasks (an artificial simple linear regression and bike, a public regression task), and one natural language processing task.
	
	In summary, our algorithm possesses the following advantages: (1) tuning-free - it requires no hyper-parameter tuning as the optimal sample-importance and method-importance are adjusted automatically at each iteration, allowing the algorithm to handle cases with sluggish convergence; (2) efficient - the computation overhead is marginal as the importance weights are computed within each forward pass and back-propagation; (3) generic - it has shown SOTA performance across various ML tasks and datasets; (4) \textit{AdaSelection} helps us better understand the proposed model and get to know how well the model is trained to prevent under/overfitting problem.
	
	\section{Related work}
	\label{rel_work}
	
	In recent years, there has been a growing interest in improving the training efficiency of DNNs by identifying the most informative sub-samples within each minibatch \cite{jiang2019accelerating,shah2020choosing,mindermann2022prioritized,sorscher2022beyond}. Training with such sub-samples can reduce the amount of data required in each iteration while maintaining the quality of the resulting model. This section gives a comprehensive review of the existing methods, analyzes their strength and weakness, and explains the technical advantage of our proposed model.
	
	\subsection{Importance sampling}
	Existing works \cite{shalev2013stochastic,johnson2013accelerating,defazio2014saga,hofmann2015variance,alain2015variance} have proved that reducing the variance of gradient is able to speed up the convergence of SGD and hence shorten the training process. Importance sampling is a common method utilized for this purpose, where the losses of certain samples are prioritized over others. The relationship between the variance of gradient estimates in SGD and the optimal sampling distribution has been established in \cite{zhao2015stochastic}. The authors determine that the sampling probability should be proportional to the gradient norm, which in simple linear classification, is proportional to the Lipschitz constant of the per-sample loss function. Another paper, \cite{katharopoulos2018not}, uses the gradient norm as a measure of sample importance and derives an upper bound for the gradient norm for optimization.

	The aforementioned methods improves the convergence speed by assigning probability to each sample when computing gradients, but still leverages all of the training samples for back-propagation. In comparison, Selective-Backprop \cite{jiang2019accelerating}, also known as the Big Loss method, further speeds up the training process by only leveraging the sub-samples with $k$ largest losses in each minibatch. In \cite{shah2020choosing}, a noiseless setting with outliers is considered and a variant of simple SGD called Small Loss is proposed. This method selects $k$ samples, then the sample with the smallest current loss, and performs and SGD-like update. The paper presents a theoretical analysis of the robustness of this approach for ML problems represented as sums of convex losses. Selective-Backprop and Small Loss employ divergent methodologies for utilizing losses, with each representing an extreme point within the spectrum of data subsampling techniques. Consequently, their efficacy is contingent upon specific situational factors, limiting their performance to particular contexts. While Another algorithm, Reducible holdout loss selection (RHO-LOSS) \cite{mindermann2022prioritized} calculates data-wise importance based on the irreducible loss difference between the hold-out dataset model and the current model. \cite{sorscher2022beyond} provides a self-supervised pruning metric with comparable performance to supervised metrics. Our work differs in that all these methods can be integrated into our policymaker to select the best candidates.
	

	\subsection{Coresets selection}
	Instead of identifying the most informative sub-samples based on sample level importances, coresets selection aims to directly select the whole sub-sample set in an optimization manner. In \cite{sener2017active}, batch active learning is defined as a coresets selection problem and batch construction is re-casted as optimizing a sparse subset approximation. Meanwhile, \cite{dong2021one} uses an approximation algorithm under the framework of Mini-batch GD, which translates the subsampling step into a combinatorial optimization problem, making it cost-efficient. However, this optimization problem must be solved with a convex optimizer, which can be time-consuming with huge variable sets, unlike the most efficient importance sampling method such as big-loss. To mitigate this problem, a Frank-Wolfe method \cite{jaggi2013revisiting} is proposed in \cite{pinsler2019bayesian} to approximate the complete data posterior of the model parameters.
	
	\subsection{Meta Learning} 
	Meta Learning is a powerful technique that allows models to fine-tune themselves based on the task at hand, offering a significant advantage over traditional machine learning methods that rely on fixed hyperparameters. This is achieved by using a bilevel optimization method, where the model's parameters are trained in such a way that a small number of gradient steps with a limited amount of training data will result in good generalization performance on the new task. One of the popular approaches to Meta Learning is the Model-Agnostic Meta-Learning (MAML) method \cite{finn2017model}, which explicitly trains the model parameters to achieve this. In our problem setting, we are similarly related to meta-learning as we aim to adaptively choose the best subsampling method to find the optimal subset for training. A related work \cite{lee2019learning} proposes a two-level optimization method that assigns weights on tasks and data samples, which could be useful in our setting as well. Another recent work \cite{baik2020meta} proposes a new weight update rule that significantly improves the fast adaptation process by using a small meta-network to generate per-step hyperparameters such as learning rate and weight decay coefficients. This meta-network could be treated as an inner loop for hyperparameter exploration in a bilevel optimization scenario, and could also be used to choose the subset in our problem setting. Overall, Meta Learning offers a promising approach for our problem as it considers the adaptive tuning of methods to fit the current task, similar to our goal of adaptively choosing the best subsampling method for training.
	
	\begin{wraptable}{r}{70mm}
		\caption{List of main notation.}
		\label{tab:list of notations}
		\resizebox{0.99\linewidth}{!}{
			\begin{tabular}{l|l}
				\hline
				Notation      & Definition                               \\ \hline
				$ i $         & Sample index          \\ \hline
				$ m $         & Candidate method index       \\ \hline
				$ t $         & Iteration index          \\ \hline
				$ f_\theta(\cdot) $         & A deep learning model       \\ \hline   
				$ l(\cdot)  $         & The loss function     \\ \hline   
				$ \alpha_{i,t}^m $         & The importance of sample $i$ at iteration $t$ given method $m$       \\ \hline
				$ w_m^t $         & The importance of method $m$ at iteration $t$   \\ \hline
				$ r_t(x_i) $         & The reward of sample $i$ at iteration $t$    \\ \hline
				$ s_{i,t} $         & The final importance of sample $i$ at iteration $t$      \\ \hline
			\end{tabular}
		}
	\end{wraptable}

	\subsection{AdaBoost} 
	
	
	AdaBoost \cite{freund2013AdaBoost} is a highly innovative ensemble learning method that was created with the purpose of increasing the efficiency of binary classifiers. This method uses an iterative approach in which it learns from the mistakes of weak classifiers and transforms them into strong ones through the assignment of weights to the data points. The algorithm is based on the principle of assigning greater weight to those data points that were misclassified in the previous iteration and reducing the weight for those that were classified correctly.
	
	To calculate the weight of each data point, we first need to define $\ell_i$ as the data-wise loss for data point $i$. AdaBoost then uses this information to determine the weight of each data point through the following:
	\begin{equation}
	\label{eq:AdaBoost}
	w_i = \frac{1}{2} \log (\frac{1+\ell_i}{1-\ell_i}),
	\end{equation}
	This equation calculates the weight $w_i$ of each data point as the log of the ratio of the sum of the loss and one minus the loss. The end result of this calculation is the AdaBoost weight, which can then be used as the importance score for each data point to facilitate the subsampling process. 
	
	\section{Proposed Method: \textit{AdaSelection}}
	\label{method}
	In this section, we will discuss our proposed method \textit{AdaSelection}, which borrows the wisdom of selected baseline methods to find the optimal subset of the batch through a majority vote. We also summarize the main notations used in our paper at Table \ref{tab:list of notations}.

	\subsection{Selective Methods}
	\label{sec:baselines}
	
	In this section, we list several baseline and state-of-the-art (SoTA) subsampling methods that we will compare with to validate the efficacy of our method. These methods include uniform, Big Loss \cite{jiang2019accelerating}, Small Loss \cite{shah2020choosing}, Gradient Norm \cite{katharopoulos2018not}, AdaBoost \cite{freund2013AdaBoost} and Coresets selection \cite{sener2017active}. Due to the complexity of solving the coresets selection problem, we develop two importance sampling methods as its approximations: (1) a mixture policy that constructs the subsampling subset with 50\% of datapoints having the biggest loss and 50\% having the smallest loss, and (2) finding datapoints closest to the mean loss of the whole batch. The intuition is, the coresets are expected to be representative enough for the whole batch, and the mean losses of the sub-samples selected using these two methods are both close to the mean loss of the whole batch.
	
	Given a mini-batch of $b$ data points and a sampling rate of $\gamma$, the key features for each baseline method are summarized as follows:
	
	\begin{itemize}
		\item \textbf{Uniform}: randomly choose $b \gamma$ data points in the whole batch;
		\item \textbf{Big Loss}: choose data samples with $b\gamma$ highest training loss;
		\item \textbf{Small Loss}: choose data samples with $b\gamma$ smallest training loss;
		\item \textbf{Gradient Norm}: choose data samples with $b\gamma$ highest norm of gradients;
		\item \textbf{AdaBoost}: choose data samples with $b\gamma$ highest weights from AdaBoost computed from \eqref{eq:AdaBoost};
		\item \textbf{Coresets Approximation 1}: choose $b\gamma / 2$ data points that have highest training loss and $b\gamma / 2$ data points with smallest training loss;
		\item \textbf{Coresets Approximation 2}: choose $b\gamma$ data points closest to average batch training loss. 
	\end{itemize}
	
	To summarize, we conclude the general framework for all our baseline methods in Algorithm \ref{alg:baseline_alg}.
	
	\begin{algorithm}[tb]
		\caption{General process for data subsampling baselines.}
		\label{alg:baseline_alg}
		\begin{algorithmic}[1]
			\STATE {\bfseries Input:} dataset $\mathcal{D}=\{(x_i,y_i)\}_{i=1}^{|\mathcal{D}|}$, batchsize $b$, sampling rate $\gamma$, loss function $\ell(\cdot)$, learning rate $\eta$, selected list $\mathcal{C}=\{\}$.
			\REPEAT
			\FOR{each epoch $e$}
			\FOR{batch $B_k$}
			\STATE  A forward pass to get losses/gradients for each data point
			\STATE  Select a subset $\mathcal{M}_k$ from $B_k$ according to a certain sub-sampling strategy 
			\STATE  Add selected data points into $\mathcal{C} = \mathcal{C} \cup \mathcal{M}_k$
			\IF{$|\mathcal{C}| = |B_k|$}
			\STATE do batch SGD update on $\mathcal{C}$
			\STATE $\mathcal{C}=\{\}$
			\ENDIF
			\ENDFOR
			\ENDFOR
			\UNTIL{Converged}
		\end{algorithmic}
	\end{algorithm}

	\begin{algorithm}[ht]
		\caption{\textit{AdaSelection} for data subsampling.}
		\label{alg:AdaSelection}
		\begin{algorithmic}[1]
			\STATE {\bfseries Input:} dataset $\mathcal{D}=\{(x_i,y_i)\}_{i=1}^{|\mathcal{D}|}$, batchsize $b$, sampling rate $\gamma$, loss function $\ell(\cdot)$, learning rate $\eta$, selected list $\mathcal{C}=None$, subsampling $M$ candidates $\{g_m\}_{m=1}^M$.
			\REPEAT
			\FOR{each epoch $e$}
			\FOR{batch $B_k$}
			\STATE  A forward pass to get data-wise losses
			\STATE  Calculate data-wise importance score $\bm{s_{i,t}}$ according to \eqref{eq:score}
			\STATE  Form a subset $\mathcal{M}_k$ from $B_k$ by selecting top highest $x_i$ with score $\bm{s_{i,t}}$
			\STATE  Add selected data points into $\mathcal{C} = \mathcal{C} \cup \mathcal{M}_k$
			\IF{$|\mathcal{C}| = |B|_k$}
			\STATE do batch SGD update on $\mathcal{C}$
			\STATE $\mathcal{C}=\{\}$
			\ENDIF
			\ENDFOR
			\ENDFOR
			\UNTIL{Converged}
		\end{algorithmic}
	\end{algorithm}
	
	\subsection{\textit{AdaSelection} for subsampling strategy}
	We now introduce our method and how it combines multiple baseline methods incorporating both the sample importance within each method and the overall importance across all the methods. 
	
	Let $f_\theta$ be a deep learning model, $l$ be the loss function, $B_t=\{(x_1,y_1), (x_2,y_2),\dots,(x_k,y_k)\}$ be the mini-batch of $k$ samples at iteration $t$, and ${g_1(\cdot),\dots, g_M(\cdot)}$ be the candidate pool of $M$ baseline methods. For a given method $g_m$, the importance weight of a sample $(x_i,y_i)$ is defined as
	\begin{equation}
	\alpha_{i,t}^m = g_m(x_i, y_i, B_t), 
	\end{equation}
	, where $g_m(x_i, y_i, B_t)$ is derived based on the loss of $(x_i, y_i)$. For example, if $m$ is the Big Loss method, then $g_m(x_i, y_i, B_t)$ will be larger on the samples with larger losses (Softmax function can be applied to achieve the purpose). 
	
	To measure the overall importance of method $m$, inspired by AdaBoost, we adjust the importance of method $m$ based on the variation of the average loss from the last iteration using:
	\begin{equation}
	w^m_t =w^m_{t-1}\times \exp\bigg({\frac{\beta|\ell^m_t-\ell^m_{t-1}|}{\ell^m_{t-1}}} \bigg),
	\end{equation}
	
	, where $\ell^m_t$ is the average loss across all the samples in the mini-batch of iteration $t$ and $\beta \in [-1,1]$. 
	
	At iteration $t$, given $\alpha_{i,t}^m$ (i.e., the importance of sample $i$ based on method $m$), and $w_t^m$ (i.e., the importance of method $m$), the overall importance of sample $i$ considering all the $M$ methods can be expressed as $\bm{s_{i,t}} = \sum\limits_{m=1}^M w^m_t\bm{\alpha_{i,t}^m}$. 
	
	Interestingly,
	the design of \textit{AdaSelection} can be viewed as a reinforcement learning (RL)/Bandit framework. In this framework, each iteration $t$ corresponds to a state represented by the training batch $B_t$, and the action taken by \textit{AdaSelection} is the selection of weights $w_t^m$ associated with a subsampling method $m$. The objective is to learn the effectiveness of each strategy $m$, reflected by the weights $w^m_t$, with the aim of identifying the optimal subsampling method that maximizes the frequency of positive rewards.
	
	There exist multiple reward functions within this context. One possible approach is to employ automated curriculum learning (CL), which rewards the selection of temporally rare samples. Incorporating CL offers several advantages. In RL/Bandit scenarios, the reward function often exhibits significant variance, leading to stochasticity in the sampling process and an unknown reward structure. By utilizing CL, it becomes possible to capture the intrinsic structure of the reward function, thus mitigating variance and enhancing stability in the learning process.
	
	As explained in \cite{bengio2009curriculum}, CL focuses on easy samples at the beginning of the training process and gradually increases the difficulty, whereas in our case, an easier sample refers to a sample with a Small Loss. To capture the intrinsic structure of the reward function and reduce variance, we design a reward function for CL in the following:
	\begin{equation}
	\label{eq:CL}
	r_t(x_i) \propto \exp\left({-{t}^{{\gamma}}\frac{\ell(i)}{\sum_{i}\ell^2(i)}}\right).
	\end{equation}
	At the start of the training process, $\frac{\ell(i)}{\sum_{i}\ell^2(i)}$ is the dominant term as $t$ is small. Consequently, the subsampling strategy will focus on small, easily learnable losses. As the dominant term $t$ increases, the reward function gradually becomes fair to all samples and has no effect.
	
	The final importance score $\bm{s_{i,t}}$ of sample $i$ at iteration $t$ is defined as:
	\begin{equation}
	\label{eq:score}
	\bm{s_{i,t}} = r_t(x_i)\sum_{m=1}^M w^m_t\bm{\alpha_{i,t}^m}
	\end{equation}
	
	

	
	By setting the threshold for selecting the most significant scores, we are able to filter the selected data points from $B$ and discard the non-selected ones. The select/not select decision is indicated by a binary variable $z_t^i$ where
	\begin{equation}
	z_t^i =\begin{cases}
	1  & \bm{s_{i,t}} \text{ above threshold} \\
	0 & \bm{s_{i,t}} \text{ below threshold}.
	\end{cases}
	\end{equation}
	
	Our training process has been completed as described in Algorithm \ref{alg:AdaSelection}.

	\section{Experiment}
	
	\begin{table*}
		\centering
		\resizebox{\columnwidth}{!}{%
			\begin{tabular}{|l|l|l|l|}
				\hline
				Dataset Name & \# of classes & \# of instances (train+test) & Configurations                                                      \\ \hline
				CIFAR10 \cite{krizhevsky2009learning}  & $10$    & $50,000+10,000$              & lr=0.01, batch=128, ResNet18                                             \\ \hline
				CIFAR100 \cite{krizhevsky2009learning}  & $100$    & $50,000+10,000$              & lr=0.01, batch=128, ResNet18                                             \\ \hline
				SVHN \cite{netzer2011reading}         & 10            & $73,257+26,032$              & lr=0.01, batch=128, ResNet18                       \\ \hline
				Simple regression ($y=2x+1$)         & NA          & $10000+5000$              & lr=0.01, batch=100, simple MLP                       \\ \hline
				Bike regression \footnote{https://www.kaggle.com/code/gauravduttakiit/bike-sharing-multiple-linear-regression}       & NA          & $730$ in total              & lr=0.01, batch=100, 2-layer MLP                       \\ \hline
				
				Wikitext-2 \cite{stephen2017pointer}      & NA          & $2,088,628 + 245,569$             & lr=0.01, batch=100, Transformer                      \\ \hline
			\end{tabular}%
		}
		\caption{Dataset descriptions and summary}
		\label{tab:dataset}
	\end{table*}
	
	\begin{figure}
		\begin{minipage}[b]{0.49\textwidth}
			\centering
			\includegraphics[width=\textwidth]{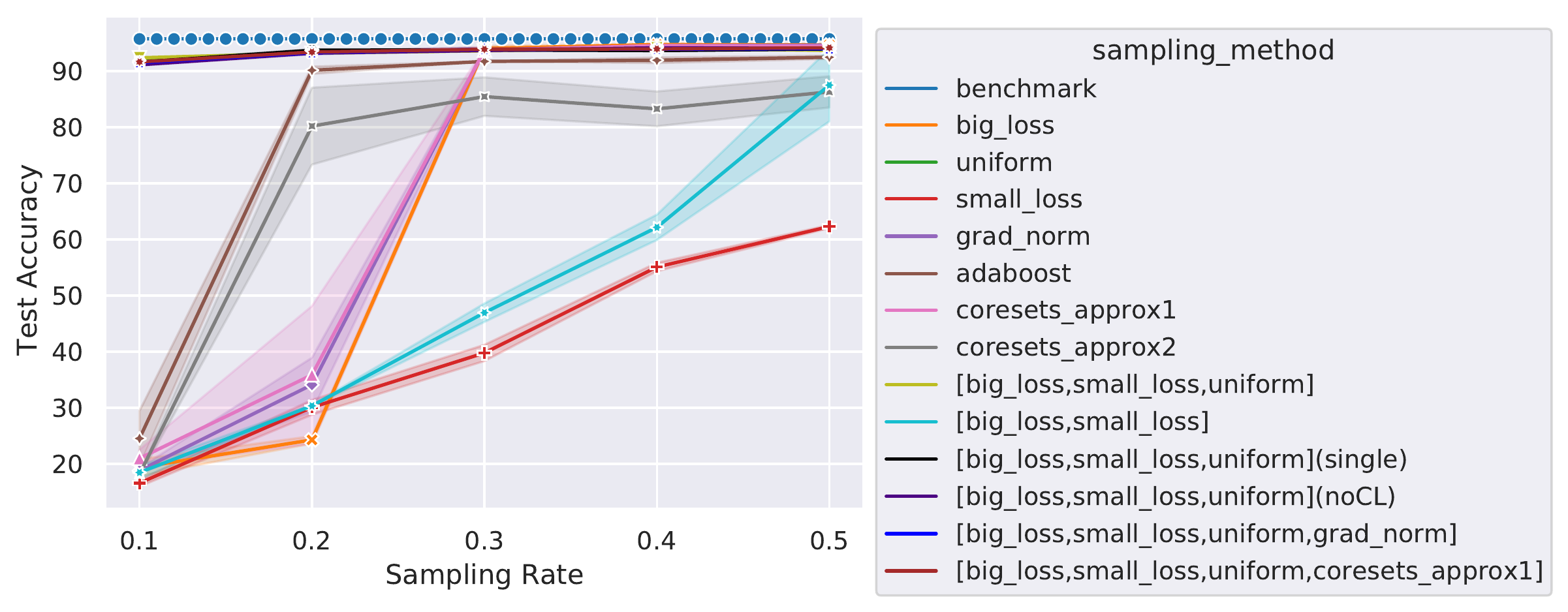}
			\caption{
				{(SVHN) Testing accuracy for different sampling rate from 0.1 to 0.5.}
			}
			\label{fig:SVHN_ACC}
		\end{minipage}%
		\hfill
		\begin{minipage}[b]{0.49\textwidth}
			\centering
			\includegraphics[width=\textwidth]{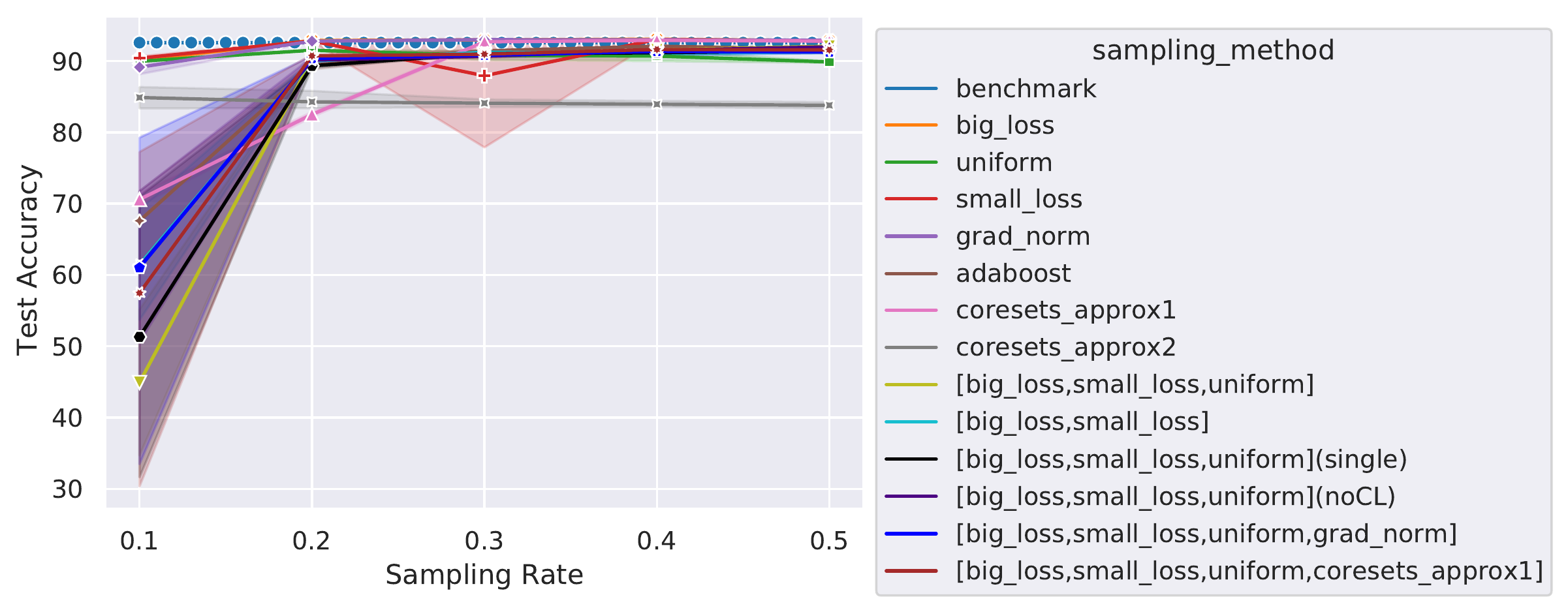}
			\caption{
				{(CIFAR10) Testing accuracy for different sampling rate from 0.1 to 0.5.}
			}
			\label{fig:Cifar10_ACC}
		\end{minipage}
	\end{figure}
	
	\begin{figure}
		\begin{minipage}[b]{0.49\textwidth}
			\centering
			\includegraphics[width=\textwidth]{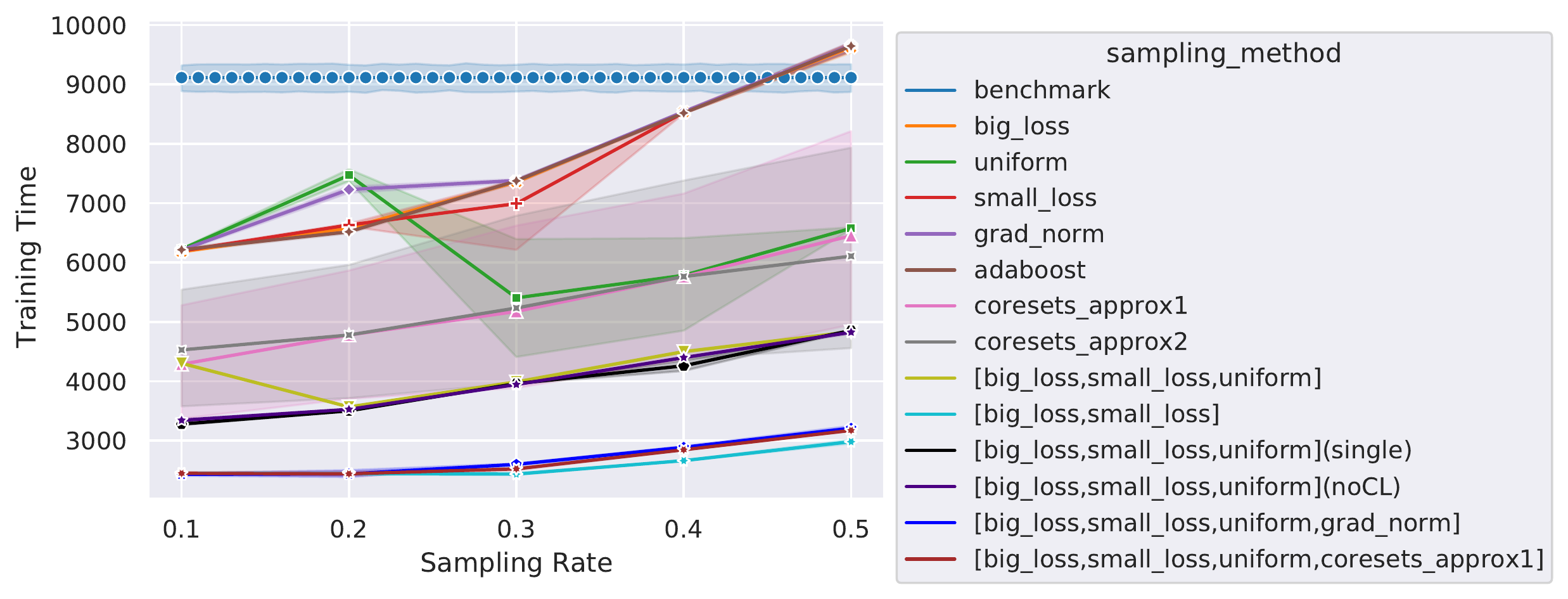}
			\caption{
				{(CIFAR10) Training Time for different sampling rate from 0.1 to 0.5.}
			}
			\label{fig:Cifar10_Time}
		\end{minipage}%
		\hfill
		\begin{minipage}[b]{0.49\textwidth}
			\centering
			\includegraphics[width=\textwidth]{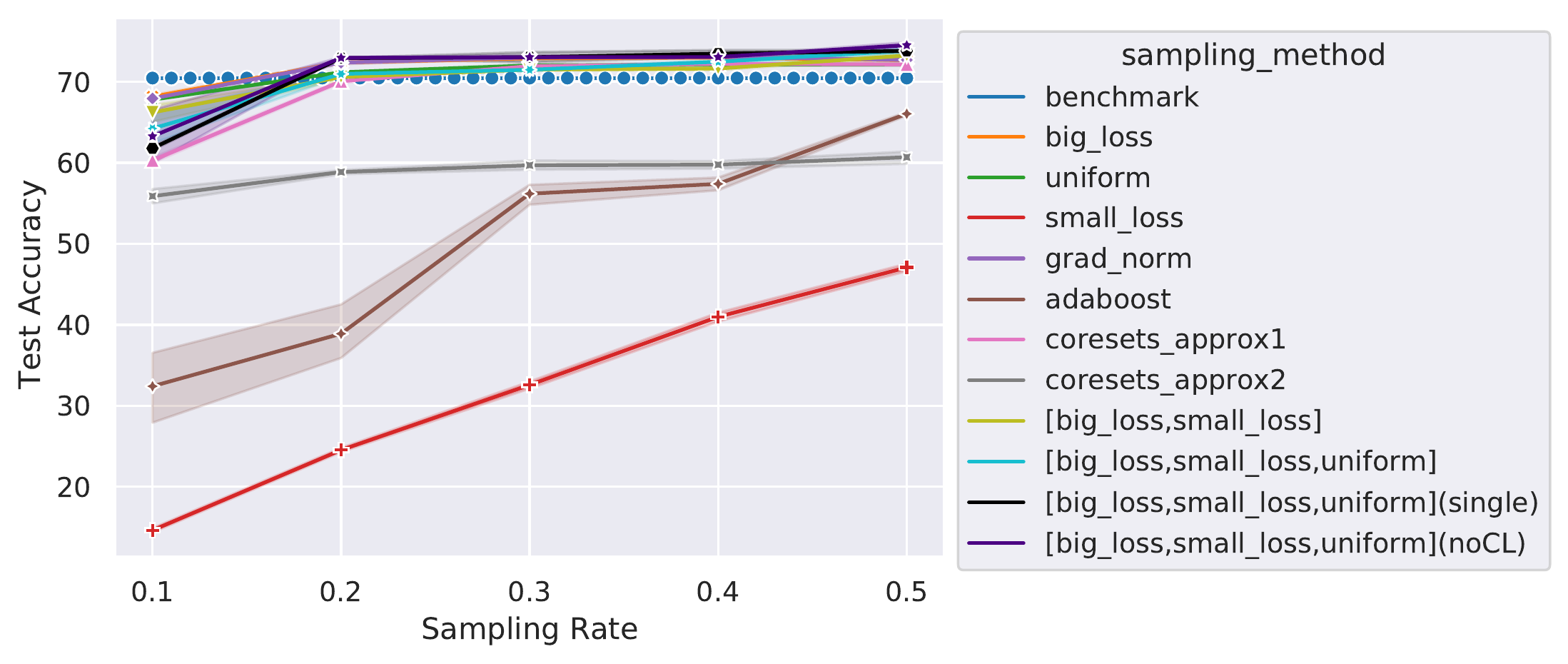}
			\caption{
				{(CIFAR100) Testing accuracy for different sampling rate from 0.1 to 0.5.}
			}
			\label{fig:Cifar100_ACC}
		\end{minipage}
	\end{figure}
	

	\begin{figure}
		\begin{minipage}[b]{0.49\textwidth}
			\centering
			\includegraphics[width=\textwidth]{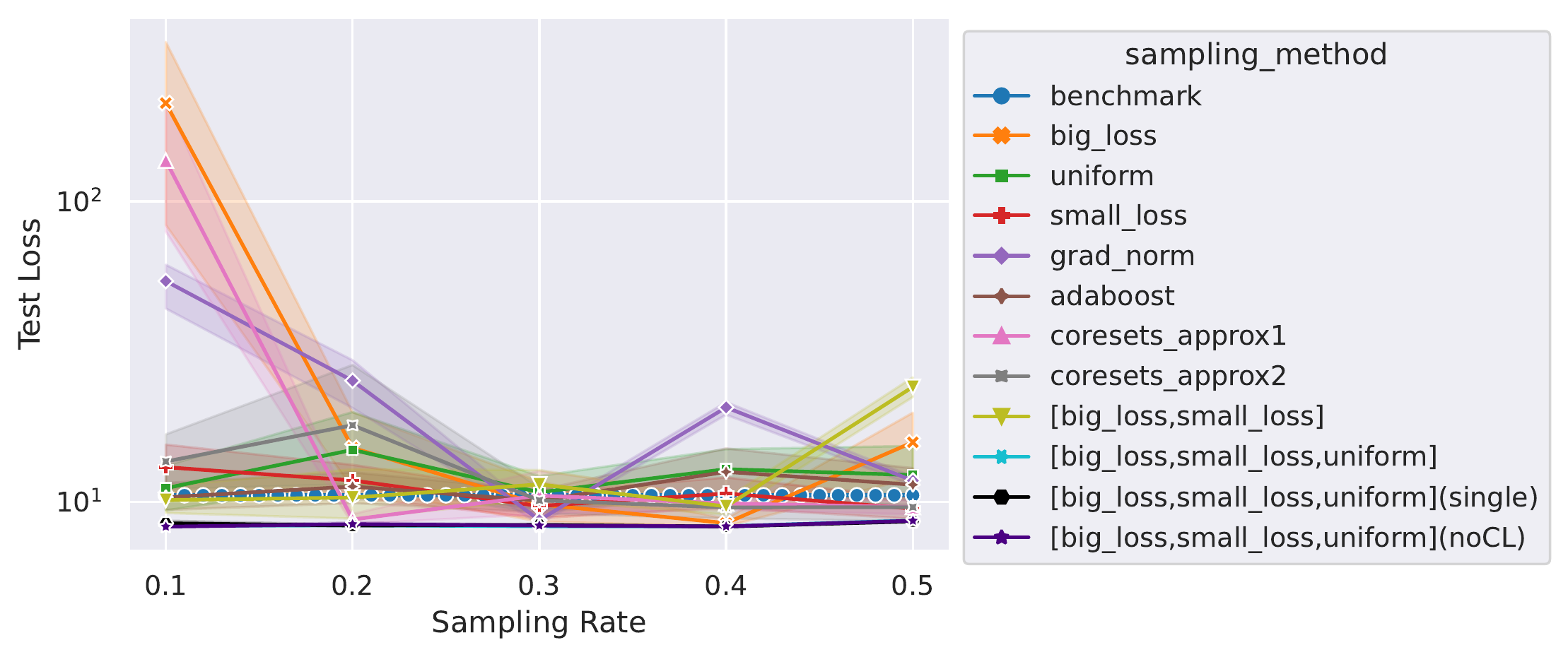}
			\caption{
				{(Regression) Testing Loss for different sampling rate from 0.1 to 0.5.}
			}
			\label{fig:regression_ACC}
		\end{minipage}%
		\hfill
		\begin{minipage}[b]{0.49\textwidth}
			\centering
			\includegraphics[width=\textwidth]{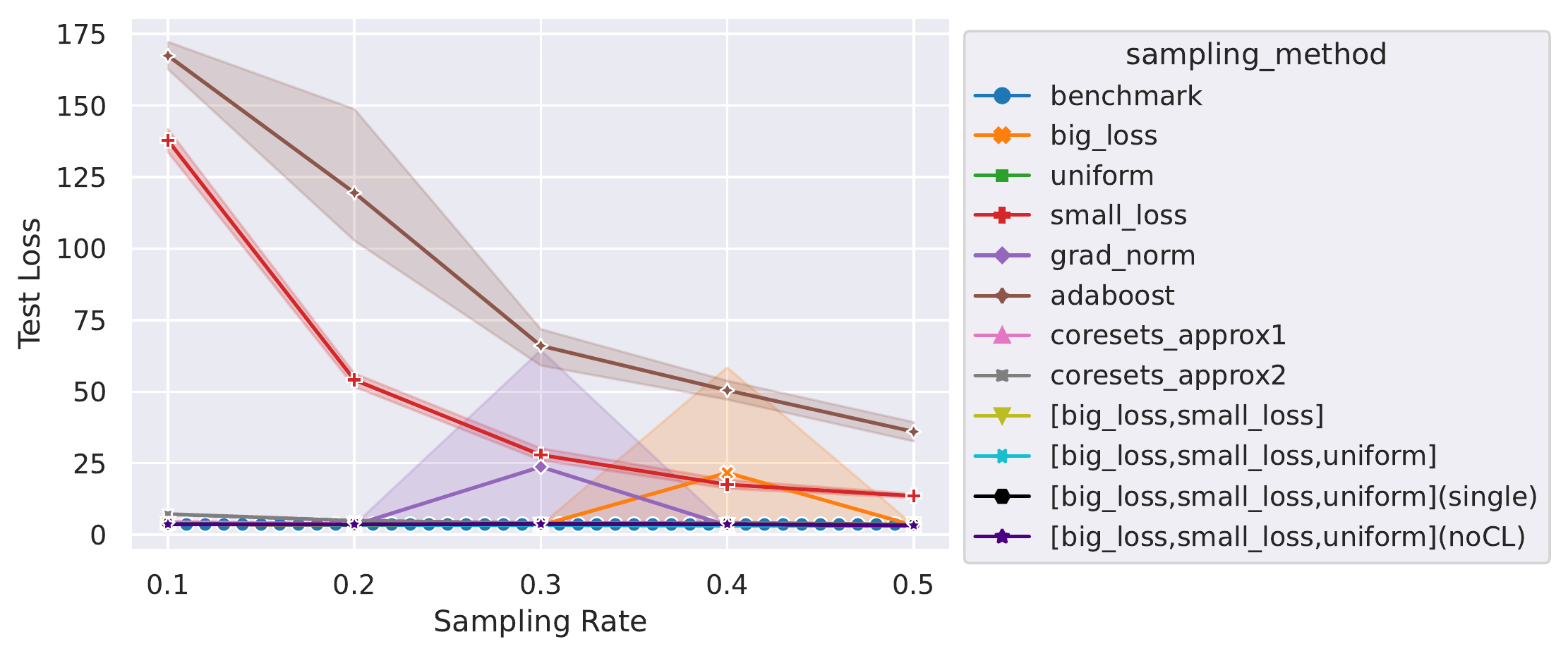}
			\caption{
				{(Regression) Testing Loss for different sampling rate from 0.1 to 0.5.}
			}
			\label{fig:Bike_ACC}
		\end{minipage}
	\end{figure}
	

	\begin{figure}
		\centering
		\subcaptionbox{SVHN dataset}{\includegraphics[width=0.33\textwidth]{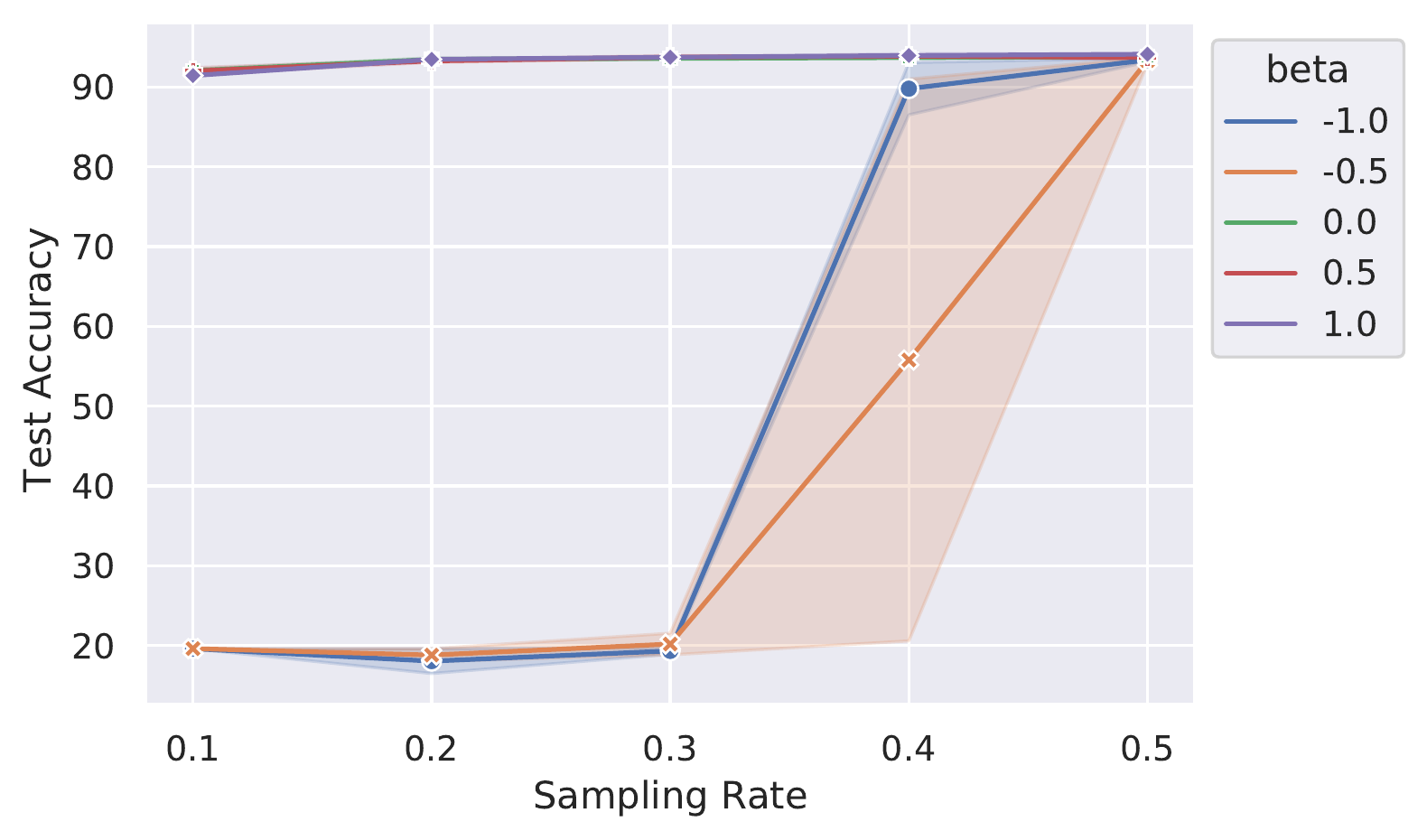}}%
		\hfill
		\subcaptionbox{Cifar10 dataset}{\includegraphics[width=0.33\textwidth]{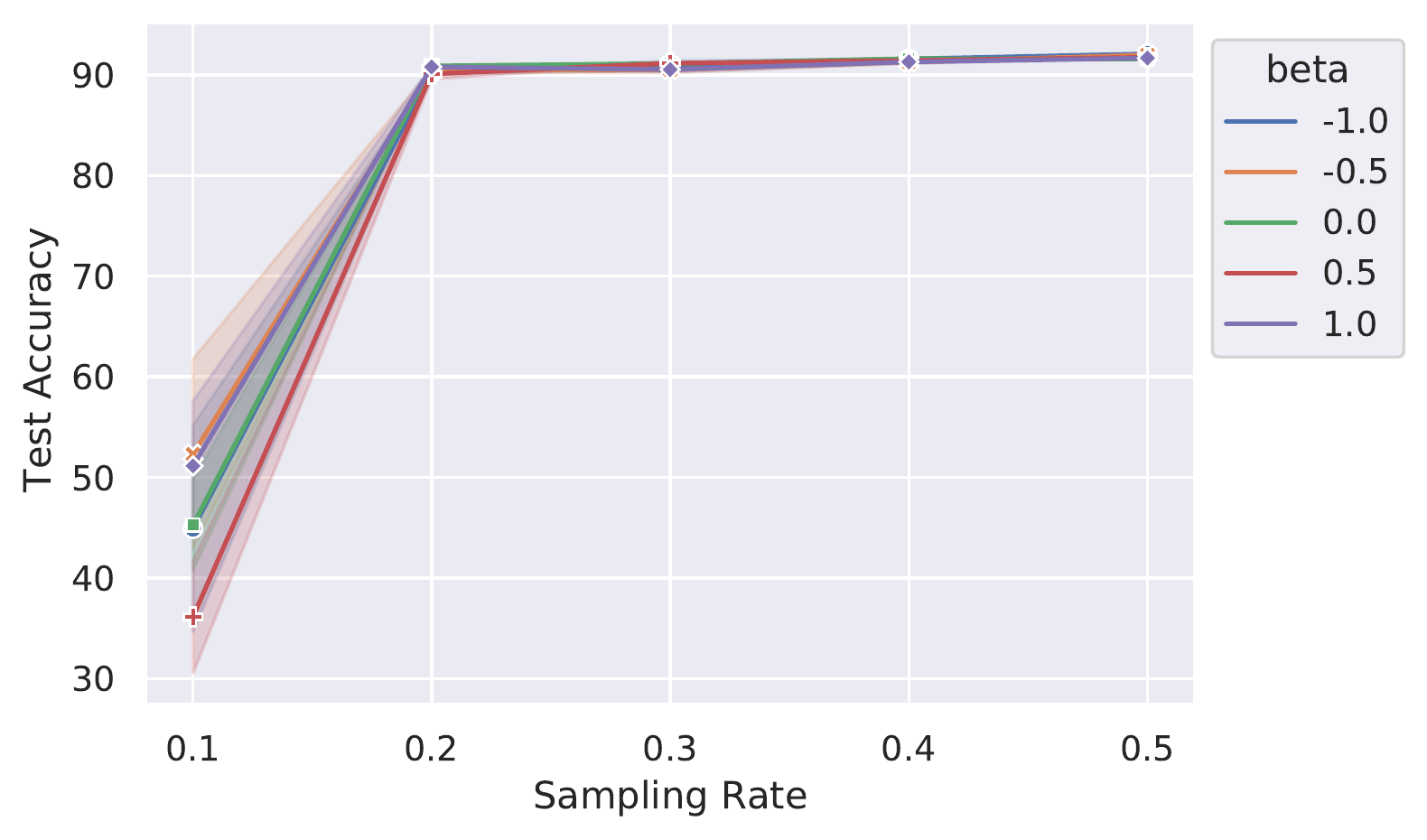}}%
		\hfill
		\subcaptionbox{Cifar100 dataset}{\includegraphics[width=0.33\textwidth]{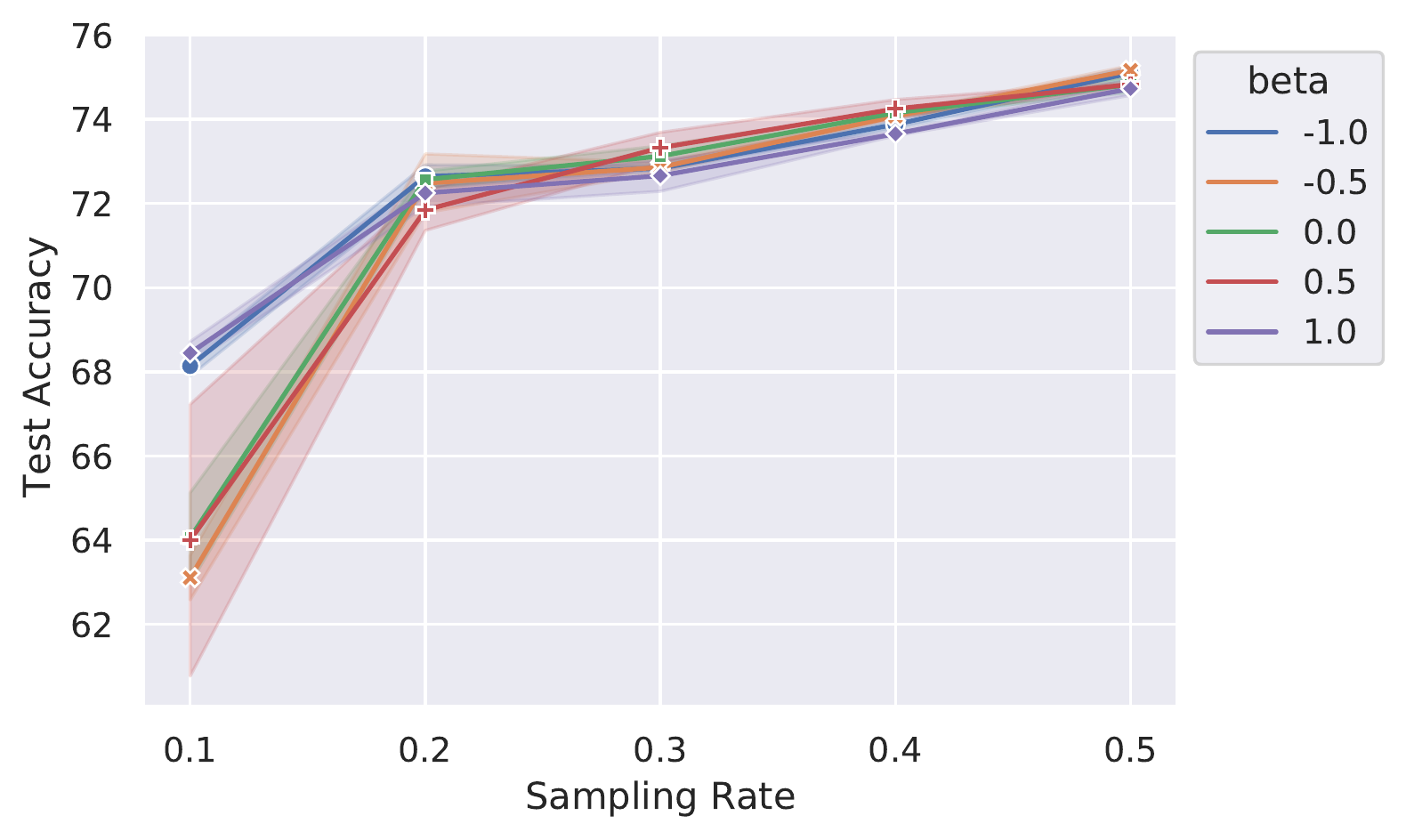}}%
		\caption{$\beta$-selection on different datasets: SVHN, Cifar10 and Cifar100. The range of $\beta$ is $[-1,-0.5,0,0.5,1]$.}
		\label{fig:beta}
	\end{figure}

	\begin{figure}
		\centering
		\subcaptionbox{SVHN dataset}{\includegraphics[width=0.5\textwidth]{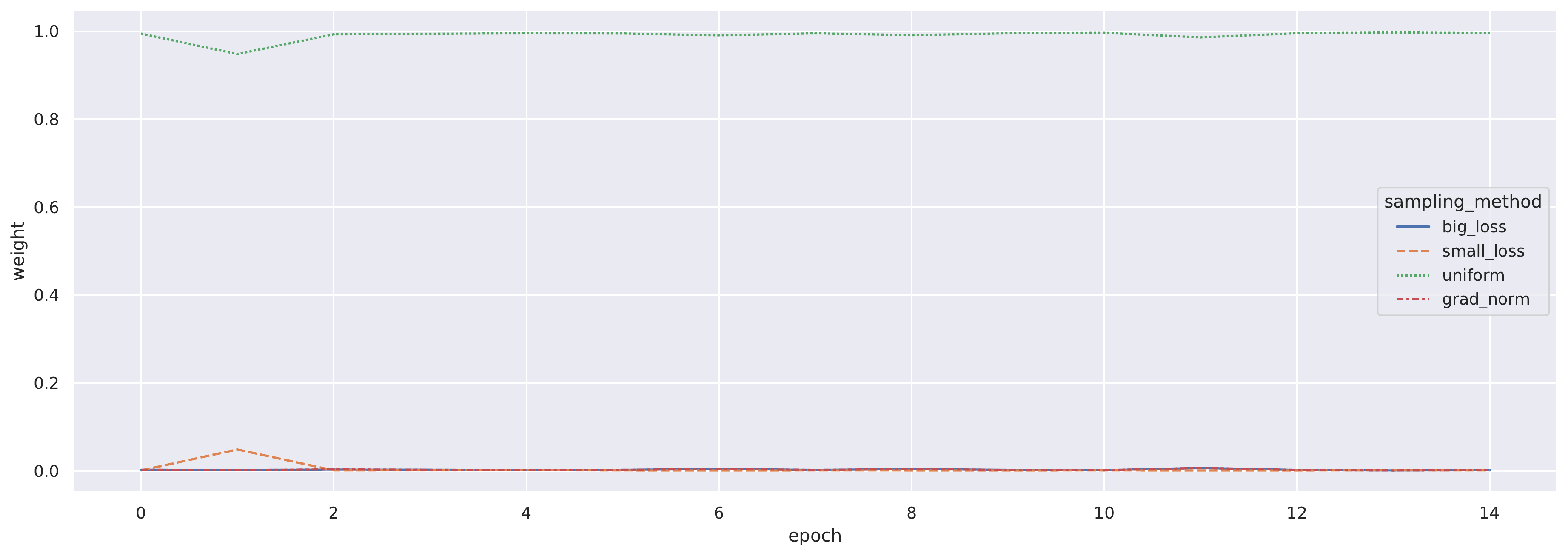}}%
		\hfill
		\subcaptionbox{Cifar10 dataset}{\includegraphics[width=0.5\textwidth]{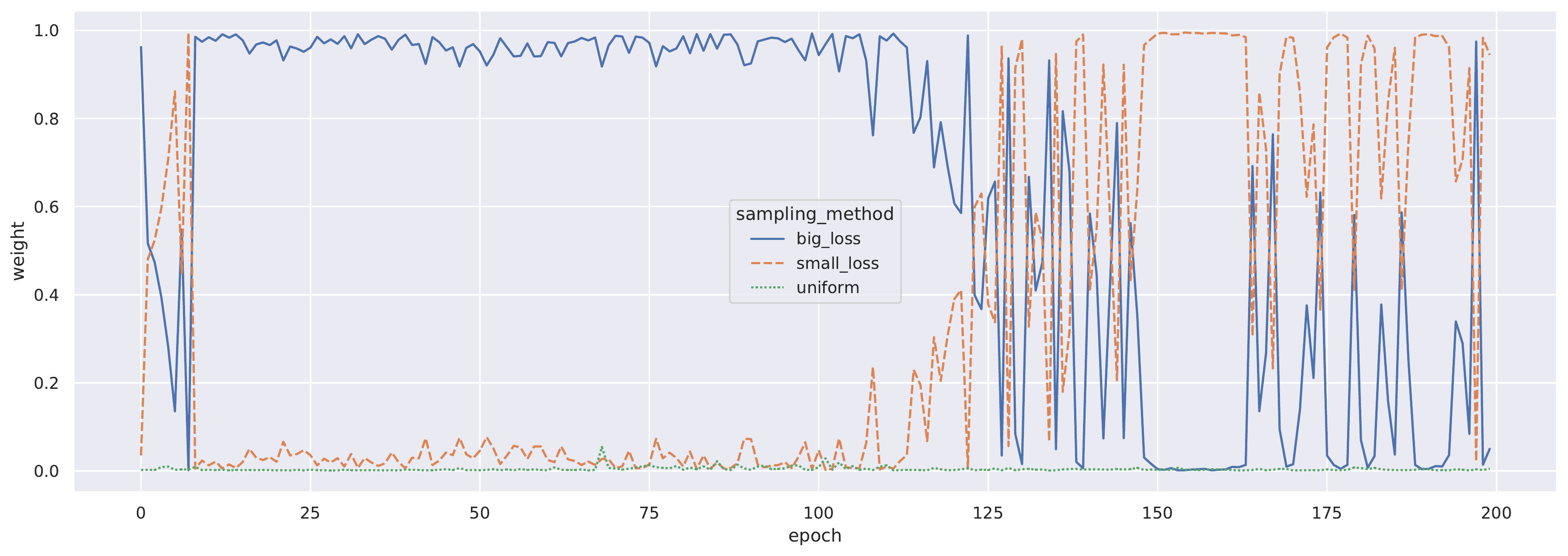}}%
		\hfill
		\subcaptionbox{Cifar100 dataset}{\includegraphics[width=0.5\textwidth]{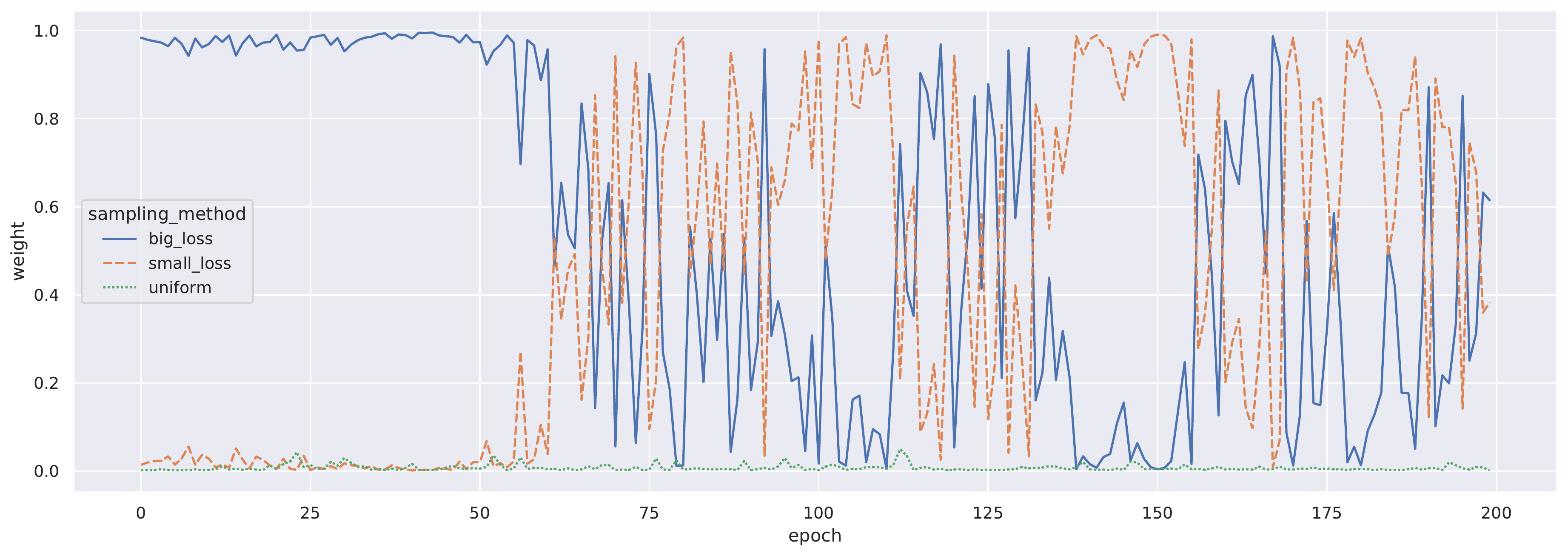}}%
		\hfill
		\subcaptionbox{Simple regression dataset}{\includegraphics[width=0.5\textwidth]{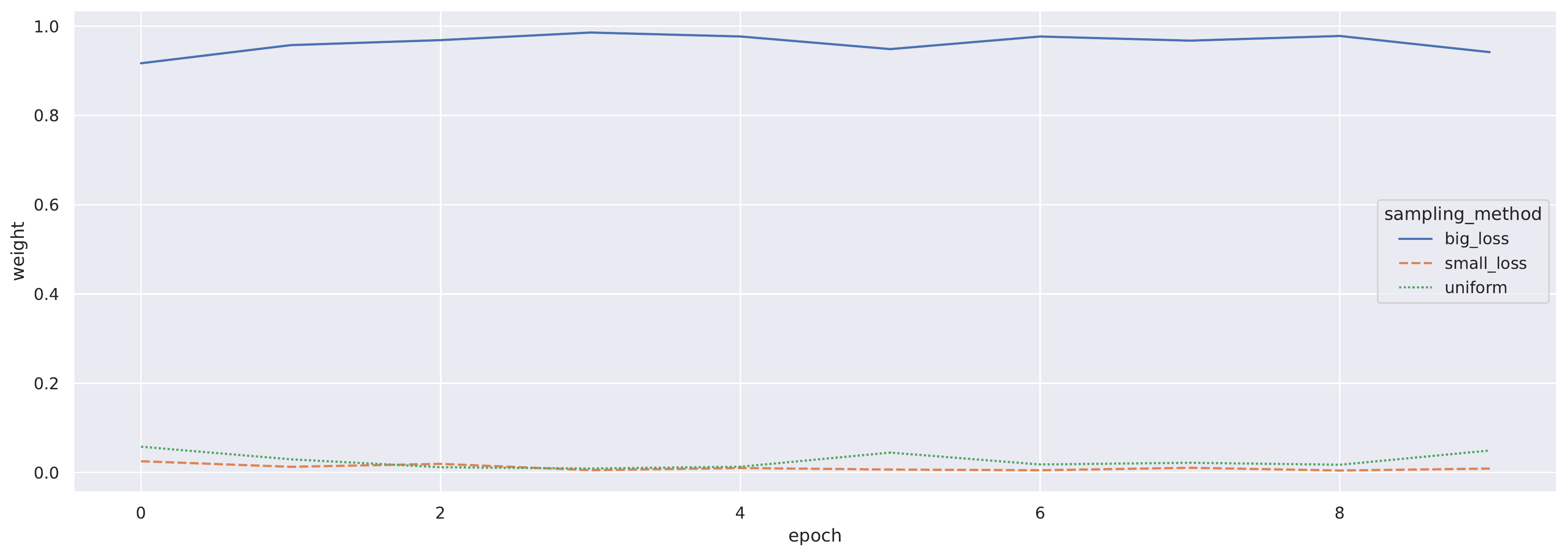}}%
		\hfill
		\subcaptionbox{Bike regression dataset}{\includegraphics[width=0.99\textwidth]{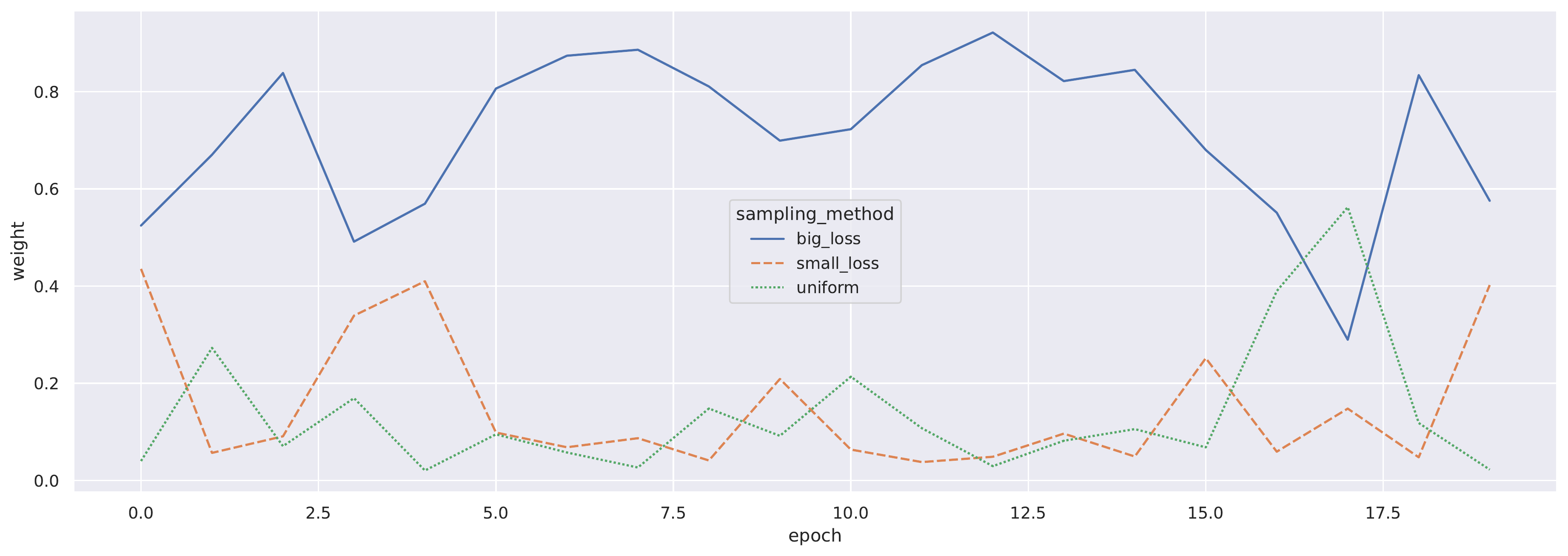}}%
		\caption{weights on candidates for \textit{AdaSelection} on different datasets: SVHN, Cifar10, Cifar100, simple regression and Bike regression tasks. The sampling rate is set to be 0.2.}
		\label{fig:weights}
	\end{figure}

	\begin{table}[ht]
		\centering
		\resizebox{\columnwidth}{!}{%
			\begin{tabular}{|l|l|l|l|l|l|l|l|l|l|}
				\hline
				& Benchmark (no sampling) & \textit{AdaSelection} & Uniform & Big Loss      & Small Loss & AdaBoost & Grad\_norm & Coresets1 & Coresets2 \\ \hline
				Cifar10    & 3.5       & 8.25         & 8.25    & \textbf{1.75} & 4.75       & 7.5      & 3.0        & 7.0              & 11.75            \\ \hline
				Cifar100   & 8.75      & \textbf{1.5} & 6.25    & 4.25          & 12.0       & 10.75    & 3.5        & 7.0              & 10.25            \\ \hline
				SVHN       & 1.0       & \textbf{4.4} & 5.8     & 6.0           & 13.8       & 9.8      & 7.2        & 6.8              & 11.8             \\ \hline
				Regression & 7.4       & \textbf{1.8} & 9.6     & 8.6           & 6.8        & 7.4      & 9.4        & 7.4              & 7.6              \\ \hline
				Bike       & 5.6       & \textbf{3.8} & 4.0     & 4.2           & 10.8       & 12.0     & 5.6        & 7.0              & 9.6              \\ \hline
				Wikitext-2       & 1.6       & \textbf{2.0}    & 2.4           & 4.4       & 7.8     & 8.2   & -     & 5.2              & 5.4              \\ \hline
			\end{tabular}%
		}
		\caption{Average ranking for testing accuracy with different sampling rate from 0.1 to 0.5. For \textit{AdaSelection}, we choose the best ranking over several choices of \textit{AdaSelection}: single choice, no CL setting, three candidates (Big Loss, Small Loss, Uniform), two candidates (Big Loss, Small Loss). Coresets1 represents Coresets approximation 1 with $50\%$ Big Loss and $50\%$ Small Loss while Coresets2 represents Coresets approximation 2 that aims to find best subset that contain datapoints that has loss that is closest to the average training loss. The bold number represents the best overall average ranking except the benchmark. }
		\label{tab:avg_ranking}
	\end{table}
	
	\begin{table}[ht]
		\centering
		\caption{Performance accuracy(classification)/loss(regression) by taking average for different sampling rate from 0.1 to 0.5. }
		\resizebox{\columnwidth}{!}{%
			\begin{tabular}{|l|l|l|l|l|l|l|l|l|l|}
				\hline
				& Benchmark (no sampling) & \textit{AdaSelection} & Uniform & Big Loss      & Small Loss & AdaBoost & Grad\_norm & Coresets1 & Coresets2 \\ \hline
				Cifar10    & 92.57\%   & 89.39\%          & 90.74\% & \textbf{92.25\%} & 91.04\%    & 85.13\%  & 91.97\%    & 84.67\%          & 84.29\%          \\ \hline
				Cifar100   & 70.47\%   & \textbf{71.38\%} & 71.05\% & 70.83\%          & 31.96\%    & 50.19\%  & 71.29\%    & 69.33\%          & 59.98\%          \\ \hline
				SVHN       & 95.71\%   & \textbf{93.38\%} & 93.32\% & 65.38\%          & 40.74\%    & 78.14\%  & 67.14\%    & 67.89\%          & 70.66\%          \\ \hline
				Regression & 10.51     & \textbf{8.42}    & 12.37   & 52.25            & 10.94      & 11.14    & 24.13      & 35.05            & 12.18            \\ \hline
				Bike       & 3.61      & \textbf{3.45}    & 3.48    & 7.08             & 50.19      & 87.90    & 7.53       & 3.75             & 7.54             \\ \hline
				Wikitext-2       & 5.45      & \textbf{5.52}    & 5.53    & 5.58             & 5.85      & 5.85    & -      & 5.61             & 5.60             \\ \hline
			\end{tabular}%
		}
		
		\label{tab:avg_acc}
	\end{table}

	In our experiments, we vary the data sub-sampling methods applied to various datasets and summarize the configurations, including the learning rate, batch size, backbone model, and dataset size in Table \ref{tab:dataset}. Our learning strategy follows the "biggest losers" approach, reducing the number of batches in each epoch. For the NN model training, we use the ResNet18 backbone for classification tasks and MLP for regression tasks, and optimize using general SGD with momentum weight decay of 0.9.
	
	We report the results for all the datasets listed in Table \ref{tab:dataset} as follows:
	
	\textbf{SVHN:} 
	To evaluate the performance of our proposed \textit{AdaSelection} method, we conduct experiments on SVHN with the benchmark, which is the original training process without subsampling. Seven methods are selected as our baselines: uniform, Big Loss, Small Loss, AdaBoost, gradient norm, coreset approximation 1, and coreset approximation 2. A detailed explanation of these methods can be found in Section \ref{sec:baselines}. Our \textit{AdaSelection} is then conducted by combining several baselines such as [Big Loss, Small Loss], [Big Loss, Small Loss, uniform], etc. The learning rate is then changed from 0.1 to 0.5 for complete training with ResNet on all tasks. Finally, we compare our proposed \textit{AdaSelection} with the top-performed baseline methods by plotting the accuracy and training time versus sampling rate to provide a complete view.
	
	
	To ensure the superiority of our proposed method, we rank its training accuracy against various methods and present the average ranking and training accuracy in Tables \ref{tab:avg_ranking} and \ref{tab:avg_acc}, respectively. The average is calculated by taking the mean of the target metric under sampling rates of 0.1, 0.2, 0.3, 0.4, and 0.5. Figure \ref{fig:SVHN_ACC} shows the testing accuracy of the SVHN method, demonstrating that with four baseline methods (Big Loss, Small Loss, uniform, grad norm/coreset1) as candidates, the training accuracy can surpass the benchmark with a $95\%$ accuracy rate even when the sampling rate is only 0.1.
	
	\textbf{Cifar10 and Cifar100:} Our analysis on two other classification tasks on Cifar10 (Figure \ref{fig:Cifar10_ACC}) and Cifar 100 (Figure \ref{fig:Cifar100_ACC}) shows that our \textit{AdaSelection} method can achieve better performance by assigning higher weights to the best performer when properly chosen candidates are used. As an example, Figure \ref{fig:Cifar10_Time} demonstrates that for the Cifar10 dataset, almost all subsampling methods reduce the training time by a minimum of 20$\%$ for all sampling rates.
	
	\textbf{Regression:} The results of the simple regression task (Figure \ref{fig:regression_ACC}) and Bike regression task (Figure \ref{fig:Bike_ACC}) show that the top performer in baseline methods for regression tasks is different than for classification tasks. The Big Loss method does not perform better than the Small Loss method in regression tasks, whereas in classification tasks, Small Loss is not performing well compared to the Big Loss method. However, the \textit{AdaSelection} method consistently selects the best performer among its candidates, resulting in the best performance among all other baselines in regression tasks.
	
\begin{figure}[ht]
		\centering
		\includegraphics[width=0.9\linewidth]{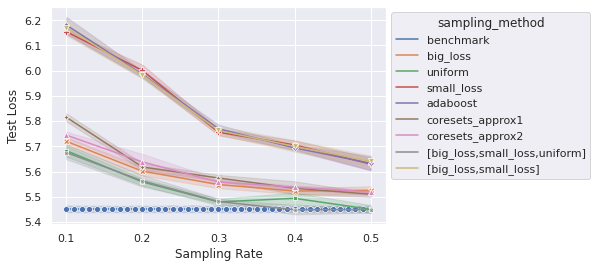}
		\caption{
			{(Transformer) Testing Loss for different sampling rate from 0.1 to 0.5.}
		}
		\label{fig:transformer_loss}
	\end{figure}

	\textbf{Transformer:} For the sequence-to-sequence language model with the Transformer module, our results show that \textit{AdaSelection} even with default parameters, outperforms other baseline methods\footnote{Gradient Norm failed to work in this transformer task because, for the NLP tasks, we cannot directly achieve gradients from the bag-of-words. Hence, we do not report its results.} as demonstrated by the results in Figure \ref{fig:transformer_loss}.
	
	The $\beta$ selection for \textit{AdaSelection} for classification tasks is shown in Figure \ref{fig:beta} to illustrate the importance of $\beta$ choice under different tasks. We can see SVHN dataset is more sensitive with a choice of $\beta$ than Cifar10/100. From Figure \ref{fig:weights}, we see the evolution of the weights for different tasks is also quite different. This suggests the necessity of our proposed methods that adaptively select the best sampling strategies during the training process.

	
	The results of the experiment demonstrate that the \textit{AdaSelection} method outperforms all the candidate baselines and is the best selection among them. Additionally, the \textit{AdaSelection} method is more robust and does not require tuning for different tasks, making it a guaranteed superior choice compared to the baseline methods.

	\section{Conclusion and Future work}
	
	
	In this paper, we introduce \textit{AdaSelection}, an adaptive subsampling method that determines the optimal policy during specific training periods. \textit{AdaSelection} combines baseline methods and assigns weights to each approach to achieve enhanced efficiency and accuracy. Experiments in image classification, text classification, linear regression, and natural language processing tasks have shown \textit{AdaSelection}'s superiority, demonstrating its effectiveness in diverse tasks.
	
	
	The future of \textit{AdaSelection} holds potential for further exploration. One possible extension is using it as an indicator for stopping the learning process, which is straightforward. \textit{AdaSelection} currently requires a forward pass for each batch to determine the training loss, which serves as the importance for feeding to baseline methods such as Big Loss or Small Loss. However, this process takes time and is not used for backward propagation, making it unnecessary. A forward pass approximation can be used instead to determine data-wise importance training, similar to a teacher-student model.

	\bibliographystyle{unsrt}
	\bibliography{ref}
	
\end{document}